\documentclass[final,5p,times,twocolumn]{elsarticle}

\usepackage{amssymb}
\usepackage{amsmath}

\usepackage[ruled,vlined,linesnumbered]{algorithm2e}
\usepackage{threeparttable}
\usepackage{booktabs}
\biboptions{sort&compress}
\usepackage[pagebackref=false,breaklinks=true,letterpaper=true,colorlinks,bookmarks=false]{hyperref}

\journal{Knowledge-Based Systems}

\begin{document}

\begin{frontmatter}



\title{Span-level Detection of AI-generated Scientific Text \\ via Contrastive Learning and Structural Calibration}


\author[label1,label2]{Zhen Yin\corref{cor1}}
\author[label1,label2,fn1]{Shenghua Wang}
\affiliation[label1]{organization={Beijing Renhe Information Technology Co., Ltd.},
            city={Beijing},
            postcode={100096}, 
            country={China}}
            
\affiliation[label2]{organization={Key Laboratory of Digital Publishing and Total Process Management of Scientific and Technical Journals},
            city={Beijing},
            postcode={100083}, 
            country={China}}

\cortext[cor1]{Corresponding author}
\fntext[fn1]{Co-first authors}

\begin{abstract}
The rapid adoption of large language models (LLMs) in scientific writing raises serious concerns regarding authorship integrity and the reliability of scholarly publications. Existing detection approaches mainly rely on document-level classification or surface-level statistical cues; however, they neglect fine-grained span localization, exhibit weak calibration, and often fail to generalize across disciplines and generators. To address these limitations, we present Sci-SpanDet, a structure-aware framework for detecting AI-generated scholarly texts. The proposed method combines section-conditioned stylistic modeling with multi-level contrastive learning to capture nuanced human–AI differences while mitigating topic dependence, thereby enhancing cross-domain robustness. In addition, it integrates BIO-CRF sequence labeling with pointer-based boundary decoding and confidence calibration to enable precise span-level detection and reliable probability estimates. Extensive experiments on a newly constructed cross-disciplinary dataset of 100,000 annotated samples generated by multiple LLM families (GPT, Qwen, DeepSeek, LLaMA) demonstrate that Sci-SpanDet achieves state-of-the-art performance, with F1(AI) of 80.17, AUROC of 92.63, and Span-F1 of 74.36. Furthermore, it shows strong resilience under adversarial rewriting and maintains balanced accuracy across IMRaD sections and diverse disciplines, substantially surpassing existing baselines. To ensure reproducibility and to foster further research on AI-generated text detection in scholarly documents, the curated dataset and source code will be publicly released upon publication.
\end{abstract}


\end{frontmatter}

\section{Introduction}
\label{introduction}

Large language models (LLMs) are rapidly permeating scholarly writing and scientific communication, delivering efficiency gains while prompting broad debate and policy updates on transparency, authorship, and responsible use, with Nature, Science, Elsevier, and Springer Nature issuing guidance that emphasizes disclosure and accountability \cite{ref1}. At the same time, hallucinations and factual inaccuracies remain frequent in reviews, abstracts, and technical exposition, further complicating quality control in academic text \cite{ref2,ref3}. These developments motivate detection solutions tailored to long, structurally organized articles, solutions that can pinpoint potentially AI-generated content at fine granularity while also providing calibrated confidence estimates to support human verification.

Most existing AIGC detectors operate either at the document level or at the paragraph level, relying on language-model perplexity, likelihood curvature under perturbations, or discriminative encoders to produce a global “AI vs. human” decision (e.g., GLTR, DetectGPT). In mixed-authorship, strongly structured scientific articles, however, such approaches rarely provide localizable and trustworthy evidence \cite{ref4,ref5,ref5-1}. Moreover, in the absence of probability calibration (for example, temperature scaling, Expected Calibration Error (ECE), or the Brier score), operating thresholds that are tuned for one venue or domain often fail to transfer reliably to another editorial workflow.

Finer-grained (sentence/token-level) methods offer improved resolution but typically do not explicitly model discourse structure. They struggle with light rewriting, long cross-paragraph substitutions, and mixed authorship, leading to over- or under-segmentation and an inability to quantify boundary uncertainty. In addition, limited use of standardized formats such as IMRaD and weak modeling of long-range dependencies encourage reliance on shortcut features (topic or terminology density), degrading generalization across generators, domains, and publication genres\cite{ref6,ref7,ref8}.

\begin{figure*}[t]
    \centering
    \includegraphics[width=0.95\linewidth]{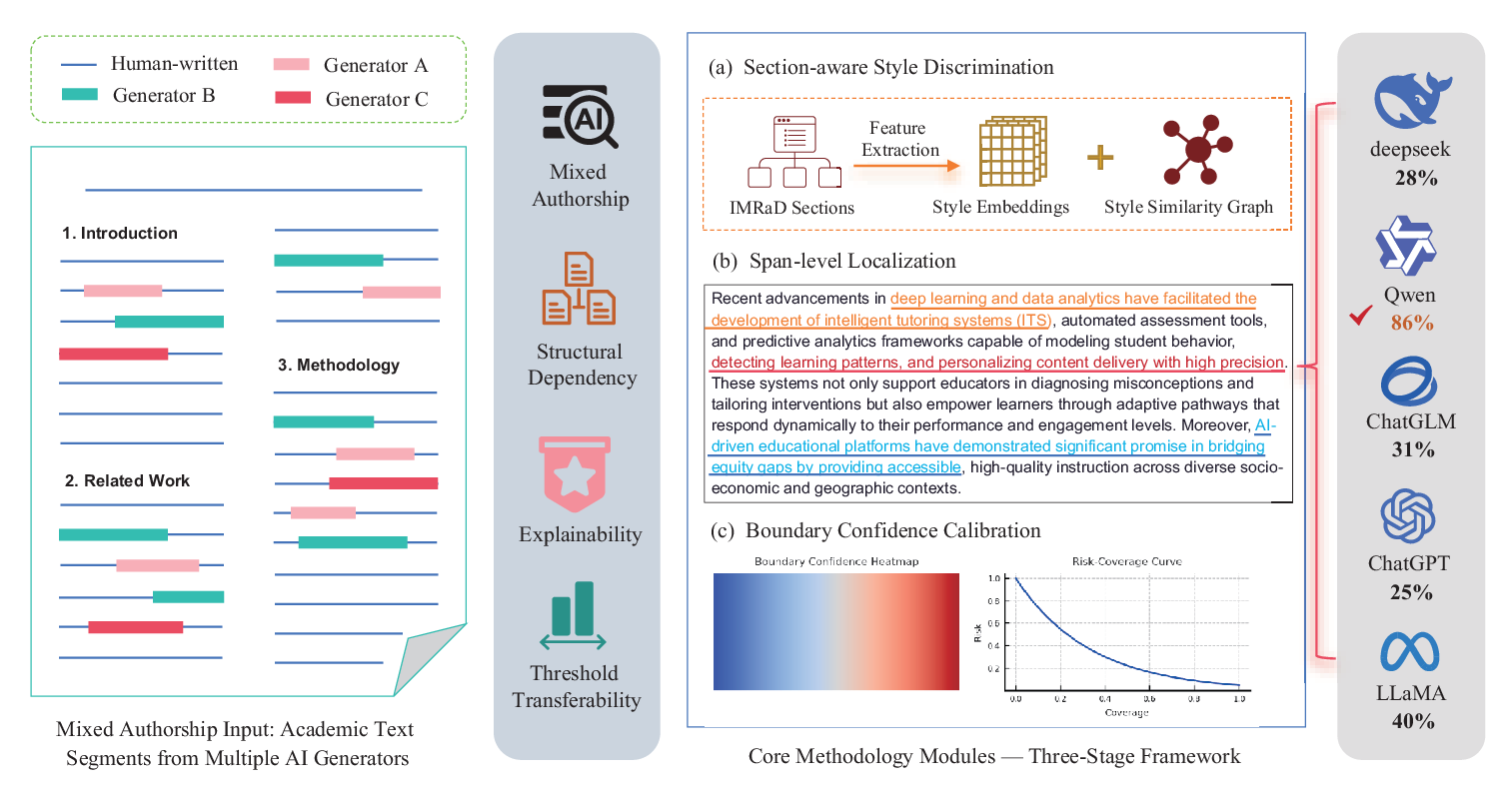}
\caption{Schematic illustration of the Sci-SpanDet framework for structure-aware detection of AI-generated content in scholarly text.}
\vspace{-.2cm}
\label{fig:Figure1}
\end{figure*}

To address these challenges, we propose Sci-SpanDet, a structure-aware detection framework for scholarly text that explicitly models micro-writing styles under section-level conditioning and jointly optimizes span-level localization with boundary calibration, thereby unifying detection–localization–calibration. Concretely, each paper is abstracted as a writing-style graph whose nodes are paragraphs and whose edges encode section membership and paragraph adjacency. We extend SimCSE-style contrastive learning to this structured setting by treating distinct IMRaD sections (e.g., \textit{Introduction}, \textit{Methods}, \textit{Results}, and \textit{Discussion}) as stylistic clusters, thereby amplifying human–AI separability within sections while mitigating topic dependence through domain-adversarial training and an information bottleneck. For localization, we couple BIO+CRF sequence labeling with a QA-style start–end pointer to jointly decode contiguous AI spans, and train a boundary-confidence predictor that enables interpretable risk–coverage control. At inference, paragraph-level posterior consistency and graph-based smoothing enhance stability and decision consistency across paragraphs. Fig.~\ref{fig:Figure1} provides an overview of the proposed framework.

The main contributions of this work are:
\begin{itemize}
    \item[\textcolor{black}{$\bullet$}] Propose a writing-style graph with a multi-level contrastive learning strategy that enhances sensitivity to section structure while remaining robust to topical semantics, thereby improving cross-generator and cross-discipline generalization in AIGC detection.
    \item[\textcolor{black}{$\bullet$}] Integrate sequence labeling and pointer networks for precise localization of AI-generated spans in mixed-authorship text, outputting calibrated confidence scores to achieve unified detection–localization with improved interpretability and practical utility.
    \item[\textcolor{black}{$\bullet$}] Enforce consistency between paragraph and token-level predictions and leverage a document-structure graph to reduce cross-paragraph label oscillation, improving stability in long-text detection.
    \item[\textcolor{black}{$\bullet$}] Construct a span-level AIGC detection dataset for scholarly articles covering multiple disciplines, generators, and mixing ratios, and define a comprehensive evaluation protocol encompassing detection, localization, and calibration.
\end{itemize}

\section{Related Work}\label{sec:rw}

\subsection{AI-generated text detection}

Early efforts in AIGC detection mainly focus on the document level, where the task is cast as binary classification at article granularity. Generative-metric approaches leverage language model statistics such as perplexity, log-likelihood, or likelihood curvature under perturbations, with representative systems including GLTR, DetectGPT, and Fast-DetectGPT \cite{ref9,ref10,ref11}. Discriminative approaches instead fine-tune pre-trained encoders to classify entire documents as human- or AI-written, occasionally enhanced with adversarial training for cross-model generalization, e.g., RADAR \cite{ref12}. These methods perform well on short open-domain text and some domain-specific benchmarks, but their limitations in scholarly long-text scenarios are evident: they provide only global decisions without locating AI-generated spans, and they are vulnerable to topical shortcuts, leading to weak cross-domain and cross-generator robustness \cite{ref13,ref14,ref14-1}.

To improve interpretability and granularity, recent studies move to span-level localization, aiming to detect AI-generated sentences or contiguous spans within mixed-authorship documents. Two major directions are sequence labeling with encoder–CRF architectures \cite{ref15,ref16} and span extraction via pointer networks or QA-style boundary regression \cite{ref17}. These approaches offer more fine-grained outputs, but still face challenges: most lack calibrated boundary confidence, hindering threshold transferability, and few exploit document structural signals such as section or paragraph context, which results in unstable segmentation under cross-paragraph or rewrite-heavy conditions \cite{ref18,ref19,ref20,ref21}. These limitations highlight the need for structure-aware, span-level detectors designed for academic long-text scenarios.

\subsection{Techniques for style and structure-aware modeling}

Stylometry and authorship attribution have demonstrated that micro-style features (function words, n-grams, syntactic patterns, rhythm) are reliable cues for distinguishing writing styles. Early shallow features, as evaluated in PAN tasks \cite{ref22}, have evolved into deep representations with contrastive learning, enabling more robust stylistic discrimination. In parallel, research on long-document modeling has introduced hierarchical attention and sparse Transformers to leverage structural cues and long-range dependencies across sections and paragraphs \cite{ref23,ref24}. These advances suggest that integrating discriminative style representations with document structure is a promising route for scholarly text detection \cite{ref25,ref26}.

Nonetheless, gaps remain when directly applying these techniques to AIGC detection. Style representations are often confounded by topical content \cite{ref27,ref28}; structure is rarely coupled with span-level localization, reducing sensitivity to discourse-specific patterns in sections such as \textit{Introduction}, \textit{Methods}, \textit{Results}, and \textit{Discussion} \cite{ref29,ref30}; and boundary predictions seldom incorporate uncertainty calibration, limiting interpretability and operational transfer. To address these issues, recent methods \cite{ref31,ref32} introduce section-conditioned contrastive learning, domain-adversarial training, and calibrated boundary modeling, improving discriminability, cross-domain generalization, and interpretability in mixed-authorship scholarly texts. Complementary to these modeling strategies, dataset and evaluation protocols emphasize diversity of generators, disciplines, and rewrite intensities, while adopting unified splits (cross-generator, cross-domain, cross-temporal) and multi-dimensional metrics, F1, AUROC for detection; Exact/Partial Span-F1 and boundary-based scores for localization; Brier score, ECE, and risk–coverage curves for calibration \cite{ref33,ref34,ref35,ref36,ref37,ref38,ref39,ref40}. Together, these advances provide the technical foundation for reliable, interpretable, and transferable AIGC detection in scholarly contexts.

\section{Proposed Method}\label{sec:method}

\noindent\textbf{Problem Formulation.}
Consider a document D with an ordered sequence of sections $\mathcal{C}=\left\{c_1,c_2,\ldots,c_K\right\}$ and a set of paragraphs $\mathcal{P}=\left\{p_1,p_2,\ldots,p_N\right\}$, where each paragraph $p_i$ belongs to section $c\left(p_i\right)$ and is tokenized as $\mathbf{x}^{\left(i\right)}=\left\{x_{1:T_i}^{\left(i\right)}\right\}$ For each token we define a latent label $y_t^{\left(i\right)}\in\left\{0,1\right\}$ (1 denotes AI-generated, 0 human-written), and we say a contiguous AI span is $s=\left(i,b,e\right)$ when $y_b^{\left(i\right)}=\cdots=y_e^{\left(i\right)}=1$. Let $\mathcal{U}\left(D\right)$ denote the document-level structural context, including neighboring paragraphs and section dependencies. The model estimates token-level posteriors $\pi_t^{(i)}=P\left(z_t^{(i)}=1 \mid D, c\left(p_i\right), \mathcal{U}(D)\right)$ and, based on these, returns the posterior field $\left\{\pi_t^{\left(i\right)}\right\}$ (optionally aggregated to sentences), an optimal set of contiguous spans $\hat{\mathcal{S}}=\arg \max _{\mathcal{S}} F\left(\mathcal{S} ;\left\{\pi_t^{(i)}\right\}\right)$ under structural and boundary-coherence constraints, and for each span a boundary confidence $q(s) \in[0,1] \approx P(s$ is a true AI-generated span $\mid D)$. The posteriors are then calibrated on a held-out set (e.g., via temperature scaling) to ensure the transferability of operating thresholds across domains and generators. Finally, we enforce paragraph–token consistency by requiring the paragraph-level posterior $P_{\mathrm{ai}}\left(p_i \mid D\right)$ to be consistent with an aggregation $h\left(\left\{\pi_t^{\left(i\right)}\right\}_{t=1}^{T_i}\right)$ (e.g., mean, max, or top-k), and we apply graph-based smoothing over the paragraph structure to yield stable, interpretable, and deployment-ready outputs for long scholarly texts.

\subsection{Overall}
\begin{figure*}[t]
    \centering
    \includegraphics[width=1\linewidth]{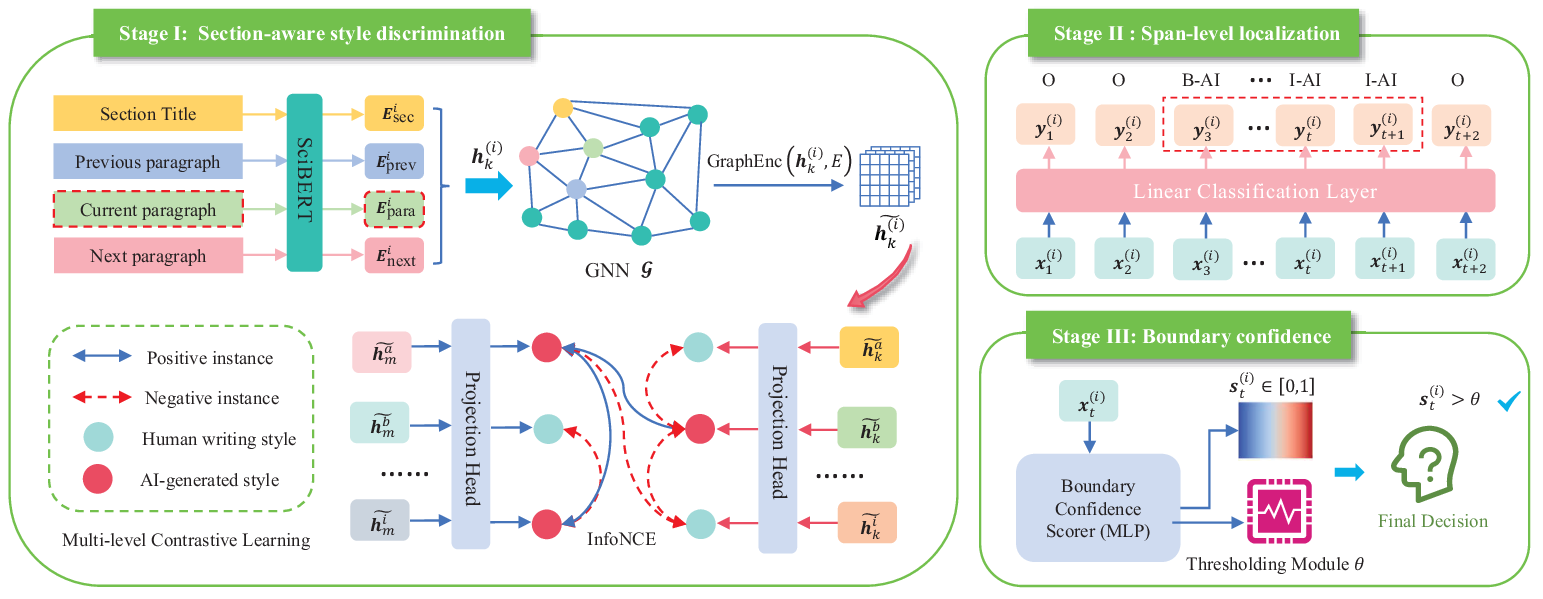}
\vspace{-.5cm}
\caption{Overview of the proposed Sci-SpanDet framework for AIGC detection, comprising section-aware style discrimination, span-level localization, and boundary confidence modeling.}
\vspace{-.3cm}
\label{fig:Figure2}
\end{figure*}

We cast AIGC detection in long scholarly documents as a unified task of detection, localization, and calibration. Given a document D with a section sequence $\mathcal{C}=\left\{c_1,c_2,\ldots,c_K\right\}$ and paragraphs $\mathcal{P}=\left\{p_1,p_2,\ldots,p_N\right\}$, the proposed Sci-SpanDet framework proceeds in three stages: first, section-conditioned micro-style modeling on a writing-style graph; second, span-level detection via BIO–CRF tagging combined with a pointer-based start–end decoder; third, boundary-confidence estimation with posterior calibration to obtain transferable operating thresholds. To enhance robustness in long documents, we additionally enforce paragraph–token consistency and apply graph-based structural smoothing. As shown in Fig.~\ref{fig:Figure2}, this three-stage process enables both fine-grained detection and stable boundary calibration, crucial for handling the complexity and length of scholarly texts.

\begin{algorithm}[t]
\SetAlCapFnt{\footnotesize}
\caption{\footnotesize{Sci-SpanDet: Training \& Inference}}
\footnotesize
\label{alg:scispandet}
\KwIn{Docs $\mathcal{D}$; epochs $E$; batch size $B$; temperature $\tau$; separation $\lambda$; cluster weight $\alpha$; min support $K$; EMA (Exponential Moving Average) momentum $\rho$; NMS (Non-Maximum Suppression) threshold $\delta$; calibration temperature $T$}
\KwOut{Calibrated span detections}

\textbf{Training:}
\For{$e=1$ \KwTo $E$}{
  \ForEach{mini-batch $\mathcal{B}\subset\mathcal{D}$}{
    \tcp{Section-aware style encoding}
    Construct fused paragraph embedding $\boldsymbol{h}_{\mathrm{para}}$; GraphEnc on $\mathcal{G}_D$ yields section-aware $\widetilde{\boldsymbol{h}}_{\mathrm{para}}$\;
    \tcp{Multi-level contrastive loss}
    Instance-level: $\mathcal{L}_{\mathrm{inst}} \leftarrow \text{InfoNCE}(\widetilde{\boldsymbol{h}}_{\mathrm{para}};\tau)$\;
    Init/refresh prototypes $\mu_{c,y}$ if $\ge K$ samples; compute \;
    Cluster-level: $\mathcal{L}_{\mathrm{clu}} \leftarrow \|\widetilde{\boldsymbol{h}}_i-\mu_{c(i),y(i)}\|^2 
- \lambda \|\mu_{c,y}-\mu_{c,y'}\|^2$\;
    \tcp{EMA prototype memory (gradient-detached)}
    $\mu_{c,y}\!\leftarrow\!(1-\rho)\mu_{c,y}+\rho\,\mathrm{mean}(\widetilde{\boldsymbol{h}}_{i\in B_{c,y}}/\|\cdot\|_2)$\;
    \tcp{Span-level localization}
    Compute CRF loss $\mathcal{L}_{\mathrm{CRF}}$ and pointer loss $\mathcal{L}_{\mathrm{ptr}}$\;
    \tcp{Compute total loss and update parameters}
    $\mathcal{L}\!=\!\mathcal{L}_{\mathrm{CRF}}+\mathcal{L}_{\mathrm{ptr}}+\mathcal{L}_{\mathrm{inst}}+\alpha\,\mathcal{L}_{\mathrm{clu}}$; update $\Theta$ by AdamW\;
  }
}
\textbf{Inference:}
Encode document $D$: build $\boldsymbol{h}_{\mathrm{para}} \!\rightarrow\! \widetilde{\boldsymbol{h}}_{\mathrm{para}}$; get BIO-CRF labels and use pointer heads 
to estimate $P_{\text{start}},P_{\text{end}}$\;
Generate candidate spans $(b,e)$; score
$s=\sum_{t=b}^{e}\log P_{\mathrm{CRF}}(I)+\phi\log P_{\text{start}}(b)+\phi\log P_{\text{end}}(e)$\;
Sort by $s$ and apply greedy NMS (IoU $<\delta$)\;
For each span $s$, compute confidence $q(s)$ by combining CRF and pointer signals; 
apply temperature scaling $q'(s)=\sigma(\log q(s)/T)$\;
\Return{\textnormal{non-overlapping spans with calibrated confidence} $q'(s)$}

\end{algorithm}

\subsection{Section-aware style discrimination} 

In long-form academic writing, stylistic characteristics are inherently conditioned by the functional roles of different sections. For example, the Introduction emphasizes research background and problem motivation, the Methods section focuses on technical details and mathematical derivations, while the Discussion centers on logical reasoning and comparative analysis of viewpoints. Relying solely on local paragraph information risks ignoring such contextual dependencies imposed by the global document structure. To address this issue, we propose a section-aware micro-style modeling approach, which jointly leverages paragraph content, surrounding context, and section-level information to achieve fine-grained characterization of writing style.

\noindent\textbf{Paragraph Representation with Contextual Fusion.}
For each paragraph, we construct three types of representations:
(i) the semantic embedding of the current paragraph $\boldsymbol{h}_{\mathrm{cur}}$;
(ii) contextual embeddings of the previous and next paragraphs $\boldsymbol{h}_{\mathrm{prev}},\ \boldsymbol{h}_{\mathrm{next}}$; and
(iii) the section-level semantic embedding $\boldsymbol{h}_{\mathrm{sec}}$ corresponding to the section in which the paragraph resides.
Rather than simple concatenation, which often introduces redundancy and excessively long inputs, we adopt a weighted fusion mechanism.
Specifically, we introduce learnable fusion coefficients $\omega_{\mathrm{cur}},\ \omega_{\mathrm{ctx}},\ \omega_{\mathrm{sec}}$ and compute the fused paragraph representation as:

\begin{equation}
\boldsymbol{h}_{\mathrm{para}}=\omega_{\mathrm{cur}} \cdot \boldsymbol{h}_{\mathrm{cur}}+\omega_{\mathrm{ctx}} \cdot \left(\boldsymbol{h}_{\mathrm{prev}}+\boldsymbol{h}_{\mathrm{next}}\right)+\omega_{\mathrm{sec}} \cdot \boldsymbol{h}_{\mathrm{sec}} \,,
\end{equation}
where $\omega_{\mathrm{cur}}+\omega_{\mathrm{ctx}}+\omega_{\mathrm{sec}}=1$, ensuring normalization and interpretability. This design allows each paragraph representation to preserve its own semantic nuances while incorporating contextual continuity and section-level stylistic information.

\noindent\textbf{Graph-based Section-aware Encoding.} We further construct a document graph $\mathcal{G}_D=\left(\mathcal{V},\mathcal{E}\right)$, where nodes correspond to paragraphs and edges include both intra-section connections and adjacency-based connections. The fused paragraph embeddings $\boldsymbol{h}_{\mathrm{para}}$ are propagated through a graph neural encoder (GAT), enabling message passing across sections and neighboring paragraphs. The final section-aware representation is given by:
\begin{equation}
{\widetilde{\boldsymbol{h}}}_{\mathrm{para}}=\mathrm{GraphEnc}\left(\boldsymbol{h}_{\mathrm{para}},\mathcal{E}\right).
\end{equation}

This mechanism ensures that each paragraph representation captures both micro-level semantic detail and macro-level stylistic context, making it more discriminative for human vs. AI authorship detection.

\noindent\textbf{Multi-level Contrastive Learning.} Upon obtaining ${\widetilde{\boldsymbol{h}}}_{\mathrm{para}}$, we introduce a multi-level contrastive objective to simultaneously capture local and global stylistic differences. 

At the instance level, paragraphs from the same section and source (human/AI) are treated as positives, while those across sources or sections serve as negatives, optimized via the InfoNCE loss:
\begin{equation}
    \mathcal{L}_{\mathrm{inst}}\ =-\log{\frac{\exp{\left(\mathrm{sim}\left({\widetilde{\boldsymbol{h}}}_i,\ {\widetilde{\boldsymbol{h}}}_i^+\right)/\tau\right)}}{\sum_{j}\exp{\left(\mathrm{sim}\left({\widetilde{\boldsymbol{h}}}_i,\ {\widetilde{\boldsymbol{h}}}_j^-\right)/\tau\right)}\ }}\ ,
\end{equation}
where $\tau$ is a temperature parameter. 

At the cluster level, we introduce section–source prototypes $\mu_{c,y}$ to capture global stylistic centers for each section $c$ and source type $y\in\left\{\mathrm{human,\ AI}\right\}$. The cluster-level contrastive loss encourages representations to stay close to their corresponding prototype while pushing apart prototypes of different sources:
\begin{equation}
   \mathcal{L}_{\mathrm{clu}}=\left\|\widetilde{\boldsymbol{h}}_i-\mu_{c(i), y(i)}\right\|^2-\lambda\left\|\mu_{c, y}-\mu_{c, y^{\prime}}\right\|^2 ,
\end{equation}
where $\lambda$ balances intra-cluster compactness and inter-cluster separation. Prototypes are dynamically updated during training to reflect the evolving distribution of representations (please refer to Appendix B for details).

The overall style-aware loss integrates both instance-level and cluster-level contrastive objectives:
\begin{equation}
\mathcal{L}_{\mathrm{style}}=\mathcal{L}_{\mathrm{inst}}+\alpha \cdot \mathcal{L}_{\mathrm{clu}} .
\end{equation}

Here, $\mathcal{L}_{\mathrm{inst}}$ enforces local stylistic discrimination between individual paragraphs, while $\mathcal{L}_{\mathrm{clu}}$ aligns paragraph embeddings with section–source prototypes to enhance global human–AI separation. The coefficient $\alpha$ controls the relative weight of prototype alignment.

\subsection{Span-level localization}

After obtaining the section-aware paragraph representation ${\widetilde{\boldsymbol{h}}}_{\mathrm{para}}$, we further model the internal token sequence of each paragraph to identify potential contiguous AI-generated spans. Unlike conventional binary classification approaches, we adopt a dual mechanism of sequence labeling and boundary prediction, which jointly yields accurate token-level assignments and stable span boundaries.

\noindent\textbf{BIO-CRF Sequence Labeling.} Each token representation $\mathbf{x}_t^{\left(i\right)}$ is fed into a Conditional Random Field (CRF) layer to produce a BIO-tag sequence (B = begin, I = inside, O = human):
\begin{equation}
    {\hat{y}}^{\left(i\right)}=\mathrm{CRF}\left(\mathbf{x}_{1:T_i}^{\left(i\right)}\right)\ .
\end{equation}

The CRF explicitly models label transitions to enforce sequence validity (e.g., an “I” cannot follow an “O” directly), thereby ensuring coherence in identifying long AI spans.

\noindent\textbf{Pointer-based Boundary Prediction.} Token-wise CRF decoding alone may lead to over-segmentation or under-segmentation. To address this, we incorporate a pointer-based boundary predictor that estimates, for each token, its probability of being a span start or end:
\begin{equation}
    P_{\mathrm{start}}\left(t\right)=\delta\left(\mathbf{w}_s^\top\mathbf{x}_t\right),\ \ P_{\mathrm{end}}\left(t\right)=\delta\left(\mathbf{w}_e^\top\mathbf{x}_t\right)\ ,
\end{equation}
where $\mathbf{w}_s,\mathbf{w}_e\in\mathbb{R}^d$ are trainable weight vectors corresponding to the start and end classifiers, respectively. Candidate spans are ranked and filtered according to these boundary probabilities, which helps mitigate fragmentary predictions and supports detection of continuous spans across sentence or even paragraph boundaries.

\noindent\textbf{Joint Decoding.} Finally, we combine BIO-CRF sequence labels with pointer-based start–end distributions through a joint scoring function:
\begin{equation}
\begin{split}
\hat{s} = \arg \max_{(b,e)} \Bigg(
   & \sum_{t=b}^e \log P_{\mathrm{CRF}}\left(y_t = I\right) \\
   & + \phi \log P_{\mathrm{start}}(b) 
   + \phi \log P_{\mathrm{end}}(e) 
\Bigg) .
\end{split}
\end{equation}
where $\left(b,e\right)$ denotes a candidate span and $\phi$ is a balancing parameter. This joint decoding mechanism harmonizes sequence consistency with boundary confidence, enabling more reliable span predictions in mixed-authorship settings.

Since CRF decoding and pointer-based boundary prediction may yield overlapping spans, we introduce a conflict resolution step to ensure coherent outputs. All candidate spans are first ranked by their joint confidence scores, after which greedy non-maximum suppression is applied: if two spans overlap beyond a predefined token-level intersection-over-union threshold $\delta$, only the higher-scoring span is retained. This procedure eliminates redundant or fragmented detections while preserving the most reliable boundaries. Furthermore, spans are constrained not to cross discontinuous sentence or paragraph boundaries, ensuring semantic and structural consistency.

\subsection{Boundary confidence and calibration} 

After span-level detection, two key challenges remain: (i) how to quantify the reliability of predicted boundaries to support human verification and risk awareness, and (ii) how to ensure that detection thresholds remain transferable across domains and generators, avoiding drastic performance drops due to distribution shift. To address these issues, we design a boundary confidence modeling and posterior calibration mechanism that enhances both interpretability and robustness of the predictions.

For each candidate span $s=\left(i,b,e\right)$, we define its confidence score by combining CRF sequence evidence and pointer-based boundary signals:
\begin{equation}
\begin{split}
q(s) = \sigma \Big(
   & \eta_1 \sum_{t=b}^{e} \log P_{\mathrm{CRF}}(y_t=I) \\
   & + \eta_2 \log P_{\mathrm{start}}(b)  + \eta_3 \log P_{\mathrm{end}}(e)
\Big)\ ,
\end{split}
\end{equation}
where $\eta_1,\eta_2,\eta_3$ are trainable weights and $\sigma\left(\cdot\right)$ is the sigmoid function. This confidence integrates internal token consistency with local boundary cues, yielding a score in $\left[0,1\right]$ that reflects the reliability of each predicted span.

Due to domain and generator variability, raw confidence estimates often suffer from over-confidence or under-confidence. We apply temperature scaling to calibrate span scores:
\begin{equation}
q^\prime\left(s\right)=\sigma\left(\frac{\log{\left(q\left(s\right)\right)}}{T}\right) .
\end{equation}
where $T>0$ is a temperature parameter learned on the validation set by minimizing calibration metrics such as Expected Calibration Error (ECE) or Brier Score. Larger $T$ values smooth overly sharp distributions, while smaller values strengthen discriminability. This adjustment ensures that detection thresholds generalize more reliably across heterogeneous test scenarios.

\section{Experiment and Analysis}\label{sec:exp} 

\subsection{Experimental setup}

\noindent\textbf{Datasets.}  We constructed a cross-disciplinary dataset comprising 100,000 annotated samples for detecting AIGC in scientific texts. Unlike existing benchmarks that primarily target generic or news-style corpora, our dataset captures the stylistic and structural complexity of scientific writing. To this end, we employed multiple LLMs, including GPT, Qwen, DeepSeek, and LLaMA, to polish, paraphrase, or rewrite human-authored scientific texts with varying degrees of modification, thereby simulating diverse real-world adversarial scenarios. Each segment was annotated at the span level to distinguish AI- from human-written content, providing fine-grained, token-level supervision that fills a critical gap in this field. This dataset enables evaluation of detection models at both the document level and the span level. For more detailed information, please refer to Appendix A.

\noindent\textbf{Baseline methods.} We compared our approach against a set of representative baselines widely used in AI-generated text detection. Specifically, we included RoBERTa-CLS \cite{ref41}, a fine-tuned supervised classifier representing the standard discriminative approach; GLTR \cite{ref4}, a statistical method based on token likelihood ranking; DetectGPT \cite{ref5}, a zero-shot detector exploiting log-probability curvature; Fast-DetectGPT \cite{ref11}, an efficiency-oriented variant of DetectGPT; DetectLLM-LRR \cite{ref42}, which leverages log-rank ratio features to enhance robustness; SeqXGPT \cite{wang2023seqxgpt}, which formulates AI-text detection as a sequence labeling problem using token-level log-probability features with convolution and self-attention networks; and PTD (Paraphrased Text Span Detection) \cite{li2024spotting}, which identifies paraphrased spans within documents to capture fine-grained rewriting behaviors. These baselines span statistical heuristics, zero-shot probability-based detectors, supervised classifiers, and fine-grained sequence labeling approaches, ensuring a comprehensive comparison with our proposed framework (Sci-SpanDet).

\noindent\textbf{Implementation details.}  Our model employs SciBERT \cite{ref43} as the backbone encoder to capture domain-specific linguistic and semantic features inherent in scientific texts. A span-level detection architecture is adopted by integrating a CRF layer with pointer-based boundary decoding for fine-grained localization of AI-generated spans. To model structural dependencies across adjacent segments, we further incorporate a graph-based section-aware encoder (GraphEnc), implemented with two GAT layers, each with 256 hidden units and a dropout rate of 0.1. The model is fine-tuned using AdamW with a learning rate of 2e-5, a batch size of 16, and a maximum input length of 512 tokens for 10 epochs. All experiments are conducted on a single NVIDIA A100 GPU with 80 GB memory.

\subsection{Main results}

Table~\ref{tab:Table1} reports the experimental results of different detection methods on our constructed academic text dataset. We evaluate models using multiple metrics: F1(AI) for the classification accuracy of AI-generated text, AUROC for overall discrimination capability, Span-F1 for fine-grained span-level localization, and ECE together with the Brier score for calibration quality of probabilistic outputs. These complementary indicators provide a comprehensive assessment of both accuracy and reliability.

\begin{table}[t]
\centering
\footnotesize  
\begin{threeparttable}
\caption{Overall detection performance on the academic text dataset}
\label{tab:Table1}
\begin{tabular}{lccccc}
\toprule
\textbf{Method} & \textbf{F1 (AI)} & \textbf{AUROC} & \textbf{Span-F1} & \textbf{ECE} & \textbf{Brier} \\
\midrule
Roberta          & 55.46 & 72.30 & ---   & 0.19 & 0.31 \\
GLTR             & 61.03 & 76.85 & ---   & 0.16 & 0.29 \\
DetectLLM-LRR    & 66.52 & 82.68 & ---   & 0.13 & 0.27 \\
DetectGPT        & 70.14 & 85.76 & ---   & 0.11 & 0.26 \\
Fast-DetectGPT   & 71.85 & 87.12 & ---   & 0.10 & 0.25 \\
SeqXGPT          & 73.42 & 88.74 & 66.25 & 0.09 & 0.24 \\
PTD	             & 74.95 & 89.36 & 68.42 & 0.09 & 0.24 \\
\textbf{Sci-SpanDet (Ours)} & \textbf{80.17} & \textbf{92.63} & \textbf{74.36} & \textbf{0.06} & \textbf{0.22} \\
\bottomrule
\end{tabular}
\vspace{-.1cm}
\end{threeparttable}
\end{table}

Overall, several clear trends emerge. First, traditional paragraph-level detectors such as RoBERTa and GLTR yield relatively low F1(AI) scores (55.46 and 61.03, respectively), confirming their limited ability to capture the stylistic and structural nuances of scientific text. Probability-based zero-shot methods, including DetectGPT and Fast-DetectGPT, substantially improve performance, achieving F1(AI) above 70 and AUROC values exceeding 85. DetectLLM-LRR also surpasses earlier heuristics by exploiting log-rank ratio features, though it remains confined to coarse-grained predictions.

Second, incorporating span-level modeling leads to further gains. SeqXGPT achieves an F1(AI) of 73.42 with a Span-F1 of 66.25, while PTD attains slightly higher overall performance (F1(AI) = 74.95, Span-F1 = 68.42). These results highlight that span-level detectors can capture localized rewriting behaviors overlooked by paragraph-level baselines, thereby offering a more fine-grained perspective on AI-assisted text.

Finally, our proposed Sci-SpanDet consistently achieves the best performance across all metrics. It obtains the highest F1(AI) of 80.17 and AUROC of 92.63, and outperforms span-level baselines with a Span-F1 of 74.36. Moreover, Sci-SpanDet yields the lowest calibration errors (ECE = 0.06, Brier = 0.22), demonstrating that it not only enhances detection accuracy and localization but also provides well-calibrated confidence estimates for reliable risk–coverage trade-offs.

In summary, the results validate the effectiveness of our structure-aware, boundary-calibrated span-level framework. By unifying detection, localization, and calibration within a single model, Sci-SpanDet substantially advances the state of the art in AI-generated text detection for scholarly writing, offering both superior accuracy and interpretability over existing baselines.

\subsection{Ablation studies and analysis}

\noindent\textbf{Module-level ablations.} To investigate the contribution of each component in our framework, we conducted a series of ablation experiments by progressively removing individual modules from Sci-SpanDet. The results are summarized in Table~\ref{tab:Table2}.

\begin{table}[t]
\centering
\footnotesize     
\begin{threeparttable}
\caption{Module-level ablation results on the academic text dataset}
\label{tab:Table2}
\begin{tabular}{lccccc}
\toprule
\textbf{Model Variant} & \textbf{F1 (AI)} & \textbf{AUROC} & \textbf{Span-F1} & \textbf{ECE} & \textbf{Brier} \\
\midrule
\quad w/o SD   & 77.92 & 90.08 & 70.21 & 0.08 & 0.25 \\
\quad w/o GraphEnc & 79.01 & 91.37 & 71.19 & 0.08 & 0.24 \\
\quad w/o MC   & 79.12 & 91.54 & 70.82 & 0.07 & 0.23 \\
\quad w/o SL   & 79.03 & 92.01 & ---   & 0.06 & 0.22 \\
\quad w/o Calibration & 80.16 & 92.62 & 74.03 & 0.12 & 0.27 \\
\quad w/o PC   & 79.48 & 92.10 & 71.14 & 0.10 & 0.26 \\
\textbf{Sci-SpanDet (All)} & \textbf{80.17} & \textbf{92.63} & \textbf{74.36} & \textbf{0.06} & \textbf{0.22} \\
\bottomrule
\end{tabular}
\begin{tablenotes}
\footnotesize
\item Abbreviations: SD = Style Discrimination, MC = Multi-level Contrastive, SL = Span-level Localization, PC = Posterior Consistency.
\end{tablenotes}
\end{threeparttable}
\end{table}

Removing the style discrimination module led to the most pronounced drop in performance, with F1(AI) decreasing from 80.17 to 77.92 and Span-F1 from 74.36 to 70.21. This highlights that modeling fine-grained stylistic cues is crucial for distinguishing between human- and AI-generated academic writing. Eliminating the graph-based encoder also resulted in noticeable degradation (F1(AI) = 79.01; Span-F1 = 71.19), demonstrating its role in capturing contextual dependencies across adjacent text segments. Similarly, removing the multi-level contrastive learning objective reduced both F1(AI) (from 80.17 to 79.12) and Span-F1 (from 74.36 to 70.82), confirming that contrastive alignment improves representation robustness against heterogeneous rewriting strategies.
We further examined the effect of removing span-level localization, where only paragraph-level predictions were retained. While overall classification remained competitive (F1(AI) = 79.03; AUROC = 92.01), Span-F1 could not be computed, underscoring the necessity of explicit boundary modeling for fine-grained detection. In addition, removing the calibration module did not substantially impact classification accuracy (F1(AI) = 80.16), but significantly deteriorated probability reliability (ECE increased from 0.06 to 0.12 and Brier from 0.22 to 0.27), suggesting that calibration is indispensable for trustworthy outputs.

Finally, we conducted an ablation on the posterior consistency constraint, which aligns span-level predictions with paragraph-level priors during inference. Removing this constraint led to a moderate decline in Span-F1 (from 74.36 to 71.25) and a clear deterioration in calibration metrics (ECE increased from 0.06 to 0.10; Brier from 0.22 to 0.26), although F1(AI) (79.48) and AUROC (92.10) remained relatively stable. This indicates that posterior consistency mainly contributes to prediction stability and cross-level interpretability, ensuring that paragraph- and span-level predictions are coherent.
In summary, the ablation studies demonstrate that each component of Sci-SpanDet contributes to its overall effectiveness. Style discrimination and span-level localization are critical for capturing the stylistic complexity of scientific texts and achieving fine-grained interpretability, while graph encoding and multi-level contrastive learning enhance contextual robustness. Moreover, calibration and posterior consistency play complementary roles in ensuring reliable, stable, and interpretable probability outputs that are essential for practical deployment in academic integrity verification.

\begin{figure}[t]
    \centering
    \includegraphics[width=1\linewidth]{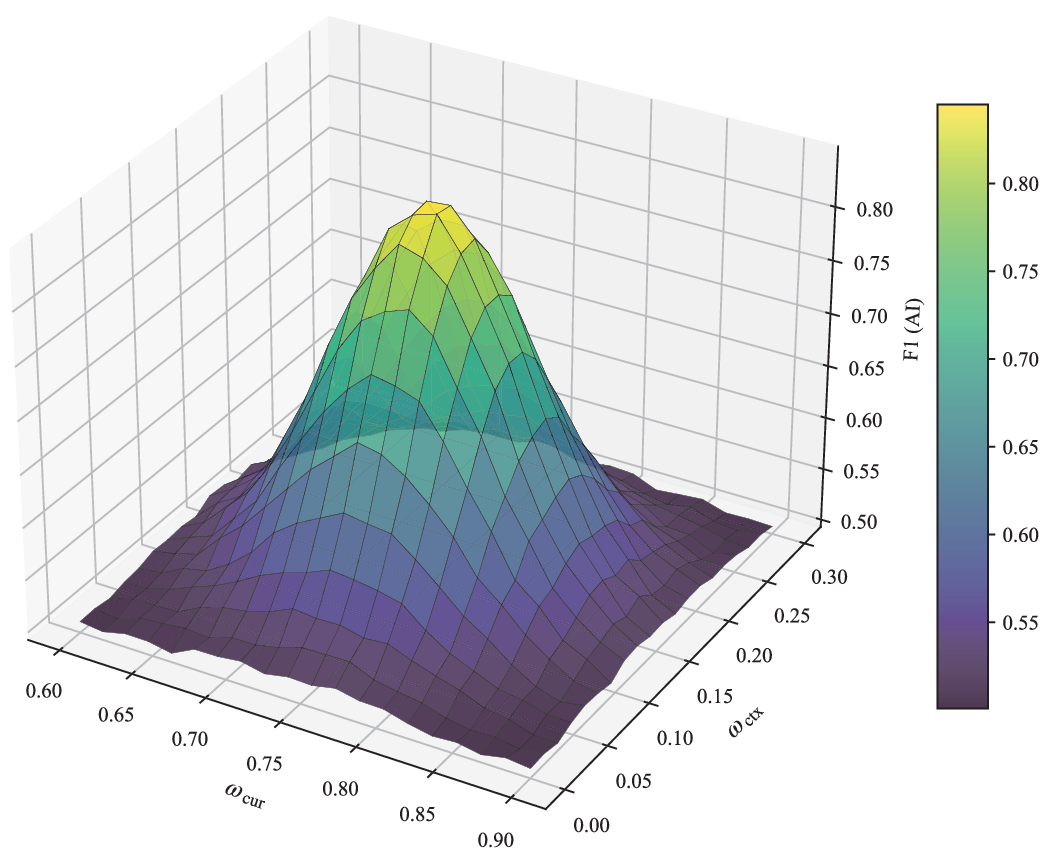}
\vspace{-.5cm}
\caption{Effect of contextual weighting on F1(AI).}
\label{fig:Figure3}
\end{figure}

\noindent\textbf{Effect of Contextual Encoding.} To further investigate the role of contextual information in detecting AI-generated academic texts, we analyze the effect of adjusting the weighting coefficients in the paragraph representation. As shown in Fig.~\ref{fig:Figure3}, F1(AI) varies with different settings of $\omega_{\mathrm{cur}}$ and $\omega_{\mathrm{ctx}}$. The model achieves the best performance when $\omega_{\mathrm{cur}}$ is set to approximately 0.75 and $\omega_{\mathrm{ctx}}$ around 0.15, highlighting that paragraph-level representation should primarily rely on the current paragraph, while a moderate amount of contextual information from adjacent paragraphs provides complementary gains. Both excessive and insufficient contextual weighting reduce detection accuracy, either by introducing noise or by ignoring useful discourse signals.

These results confirm that balancing local and contextual information is crucial for effective detection of AI-generated academic texts. Optimal weighting enables the model to emphasize the semantic fidelity of the current paragraph while still leveraging broader discourse cues.

\begin{figure}[t]
    \centering
    \includegraphics[width=0.9\linewidth]{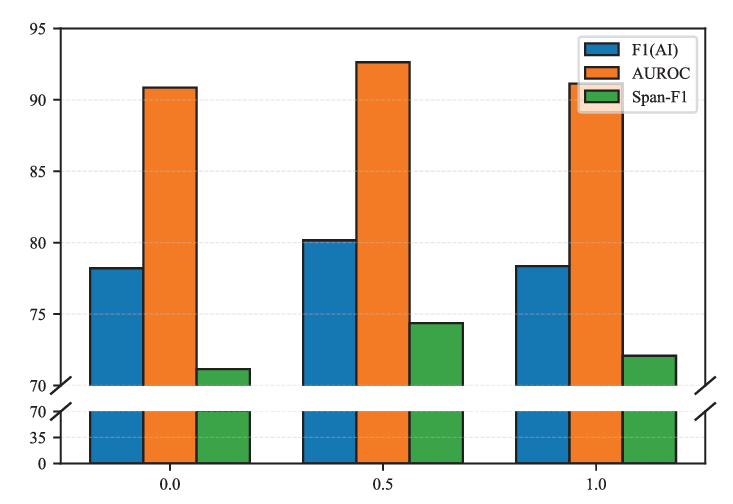}
\vspace{-.2cm}
\caption{Effect of the balancing coefficient $\alpha$ in the style-aware loss.}
\label{fig:Figure4}
\end{figure}

\noindent\textbf{Effect of the Balancing Coefficient in Style-Aware Loss.} We further examined the impact of the balancing coefficient $\alpha$, which controls the relative contribution of cluster-level contrastive alignment. Three configurations were tested: $\alpha=0, 0.5, 1.0$.
As shown in Fig.~\ref{fig:Figure4}, the results exhibit clear differences across settings. When $\alpha=0$, the model reduces to instance-level contrastive learning only, yielding F1(AI)=78.21, AUROC=90.85, and Span-F1=71.15. This configuration struggles to capture global stylistic consistency, resulting in weaker localization accuracy. Increasing $\alpha$ to 0.5 significantly improves performance, with F1(AI)=80.17, AUROC=92.63, and Span-F1=74.36, indicating that moderate prototype alignment provides an effective complement to local discrimination. However, assigning equal weight to cluster-level alignment ($\alpha=1.0$) leads to a drop in performance (F1(AI)=78.36, AUROC=91.14, Span-F1=72.08), suggesting that excessive emphasis on global prototypes suppresses fine-grained stylistic cues.
Overall, these findings demonstrate that a balanced setting is essential. The best results are obtained with $\alpha=0.5$, which achieves the highest performance across all three metrics, confirming that moderate cluster-level alignment enhances robustness without over-regularization.

\noindent\textbf{Effect of GraphEnc configurations.} We evaluate GraphEnc under different architectural configurations by varying the number of layers (1/2/3), hidden dimensions (128/256/512), and dropout rates (0/0.1/0.3). Rather than performing an exhaustive $3\times3\times3$ grid search, we conduct one-factor scans around the default setting (2 layers, 256 hidden, dropout = 0.1) and supplement them with a few two-way combinations (Layers×Dropout, Layers×Hidden) to probe local interactions. The evaluated configurations are listed in Table~\ref{tab:Table3}.

\begin{table}[t]
\centering
\footnotesize
\begin{threeparttable}
\caption{Effect of varying GraphEnc configurations on the academic text dataset}
\label{tab:Table3}
\begin{tabular}{lccccc}
\toprule
\textbf{Layers} & \textbf{Hidden Dim} & \textbf{Dropout} & \textbf{F1 (AI)} & \textbf{AUROC} & \textbf{Span-F1} \\
\midrule
1 & 256 & 0.1 & 78.43 & 90.82 & 70.25 \\
\textbf{2} (default) & \textbf{256} & \textbf{0.1} & \textbf{80.17} & \textbf{92.63} & \textbf{74.36} \\
3 & 256 & 0.1 & 79.08 & 91.46 & 71.54 \\
\cmidrule(lr){1-6}
2 & 128 & 0.1 & 78.75 & 91.02 & 70.88 \\
2 & 512 & 0.1 & 79.01 & 91.35 & 71.50 \\
2 & 256 & 0.0 & 78.92 & 90.95 & 71.02 \\
2 & 256 & 0.3 & 78.84 & 91.08 & 71.07 \\
\bottomrule
\end{tabular}
\end{threeparttable}
\end{table}

Results show that the default configuration consistently achieves the best performance (F1(AI) = 80.17, AUROC = 92.63, Span-F1 = 74.36). A shallower single-layer encoder substantially reduces Span-F1 (70.25), while a deeper three-layer design offers no improvement and even decreases AUROC, suggesting over-smoothing. Similarly, using either a smaller (128) or larger (512) hidden dimension degrades performance compared to the balanced 256-dimension setup. Dropout analysis confirms that a moderate rate (0.1) is optimal, as removing dropout (0.0) or applying stronger regularization (0.3) leads to less stable predictions.

Overall, these results demonstrate that a moderately deep encoder with balanced capacity and regularization, configured as a two-layer network with 256 hidden units and a dropout rate of 0.1, provides the best trade-off and ensures reliable span-level detection of AI-generated content in scientific texts.

\noindent\textbf{Effect of Rewriting Intensity.} To evaluate the robustness of different detection methods under varying degrees of text modification, we simulate four levels of rewriting intensity: \textit{Light} (0–10\%), \textit{Medium} (10–20\%), \textit{Heavy} (20–30\%), and \textit{Extreme} (greater than 30\%). These settings represent progressively adversarial scenarios, ranging from minor lexical substitutions to substantial sentence restructuring, thereby providing a systematic assessment of model resilience against rewriting attacks.

\begin{figure}[t]
\centering
\includegraphics[width=0.95\linewidth]{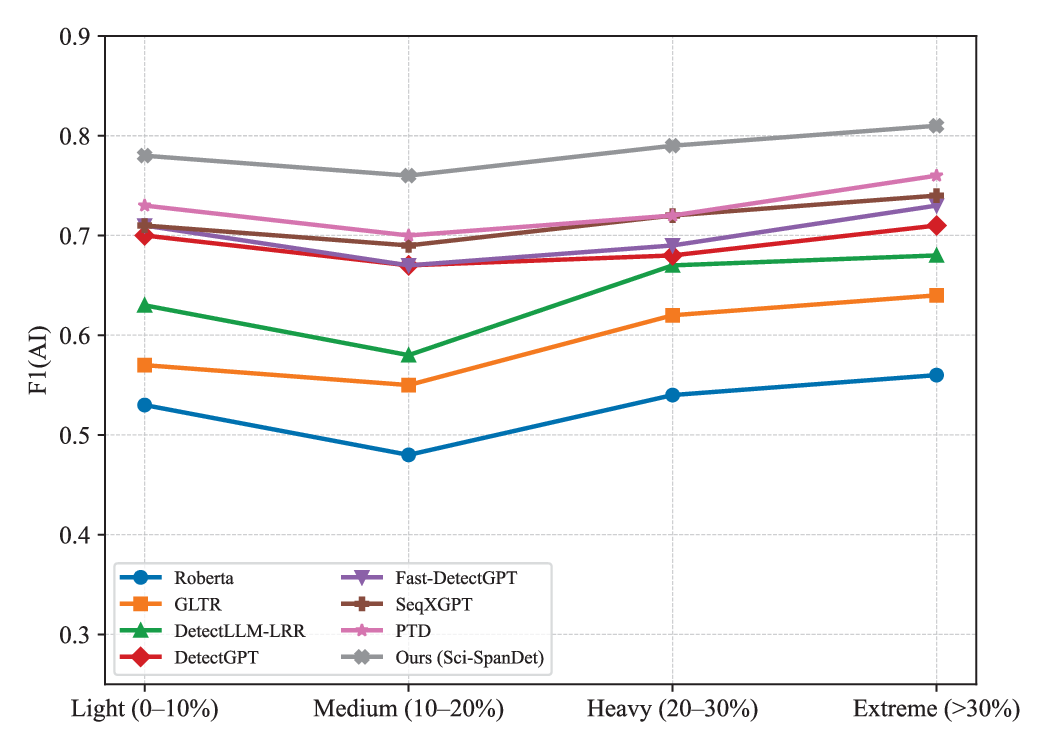}
\vspace{-.2cm}
\caption{Effect of rewriting intensity on detection performance.}
\label{fig:Figure5}
\end{figure}

As shown in Fig.~\ref{fig:Figure5}, different detectors exhibit distinct sensitivities to rewriting intensity. Traditional supervised and statistical models such as RoBERTa and GLTR show the largest degradation, with F1(AI) dropping to 0.48 and 0.55 under medium rewriting, reflecting their reliance on surface lexical cues. Zero-shot approaches including DetectGPT, Fast-DetectGPT, and DetectLLM-LRR perform more robustly, maintaining F1(AI) above 0.67 across all levels, though moderate declines still occur under medium perturbation.  

Span-level methods demonstrate clear advantages in capturing localized rewriting. SeqXGPT and PTD consistently outperform paragraph-level baselines, reaching F1(AI) up to 0.74 and 0.76, respectively. Nonetheless, our proposed Sci-SpanDet achieves the highest and most stable performance, with F1(AI) ranging from 0.78 to 0.81 across all intensities. Its robustness, particularly under medium and heavy rewriting where other methods falter, derives from boundary-aware span supervision and section-conditioned style modeling, enabling more reliable detection of AI-generated academic text.

\noindent\textbf{Embedding Visualization and Separability Analysis.} To better understand the relative separability between human-written texts and outputs from different LLM families, we visualize their embeddings in a two-dimensional space. This analysis highlights how newer model versions tend to produce texts that are stylistically closer to human writing, while earlier versions remain more easily distinguishable.

\begin{figure}[t]
    \centering  
    \includegraphics[width=1\linewidth]{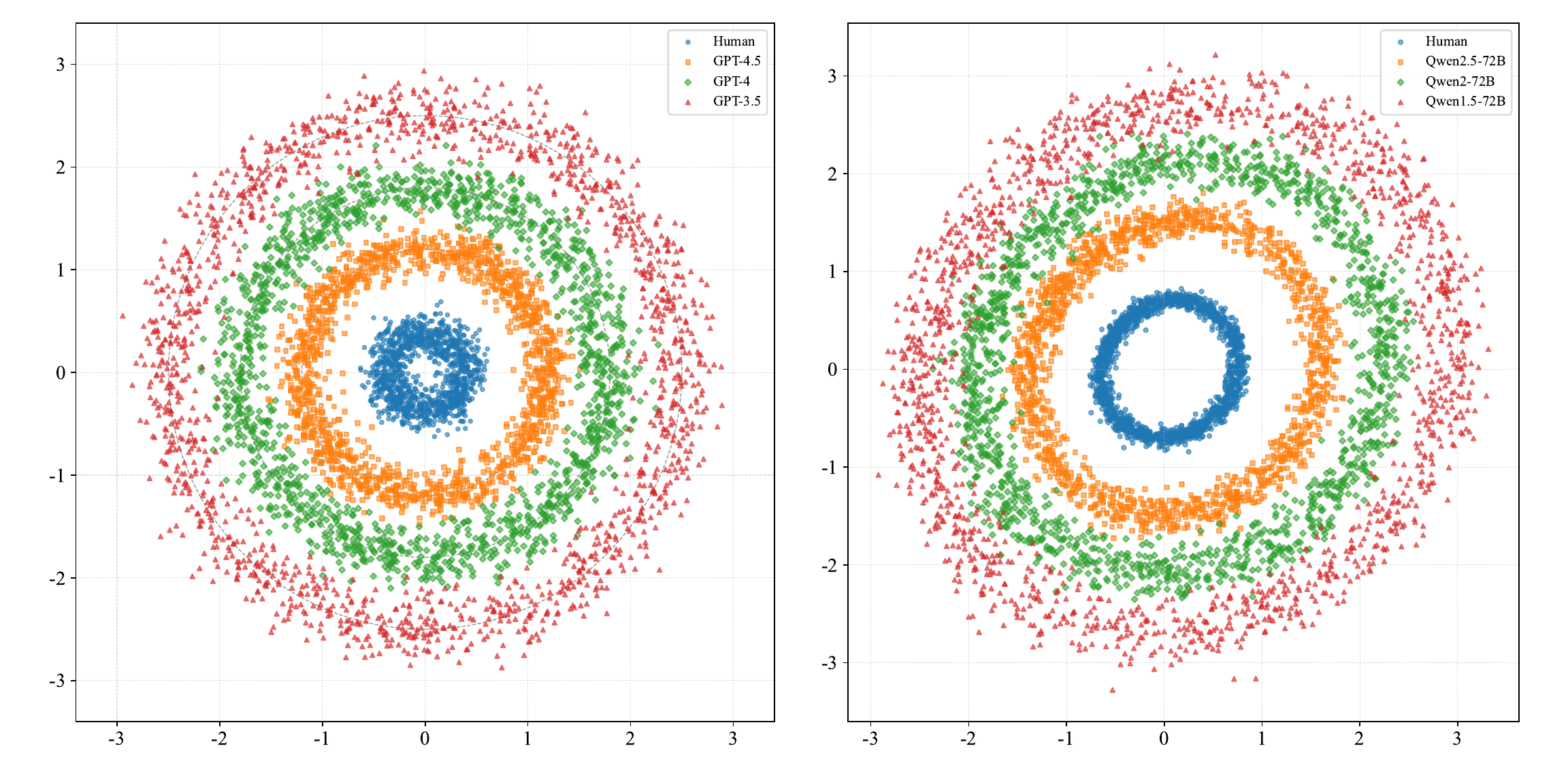}
\vspace{-.3cm}
\caption{Embedding visualization of Human vs GPT and Qwen families.}
\label{fig:Figure6}
\end{figure}

Fig.~\ref{fig:Figure6} illustrates the relative embedding distributions of Human text and different LLM-generated versions. On the left, Human samples form the innermost cluster with the smallest radius and highest density. Surrounding layers correspond to GPT-4.5, GPT-4, and GPT-3.5, respectively, where older versions exhibit progressively larger radii, indicating greater separability from Human writing. On the right, a similar pattern is observed for the Qwen family, where Human texts remain at the center and Qwen2.5-72B, Qwen2-72B, and Qwen1.5-72B form concentric outer layers.

The radius of each ring reflects the relative distance to Human representations: larger radii suggest higher separability and easier detection, while smaller radii imply closer stylistic alignment and thus more challenging detection. Moreover, the random perturbations in the distributions prevent them from forming idealized circles, making the visualization closer to realistic embedding dispersion.

\begin{figure}[t]
    \centering
    \includegraphics[width=1\linewidth]{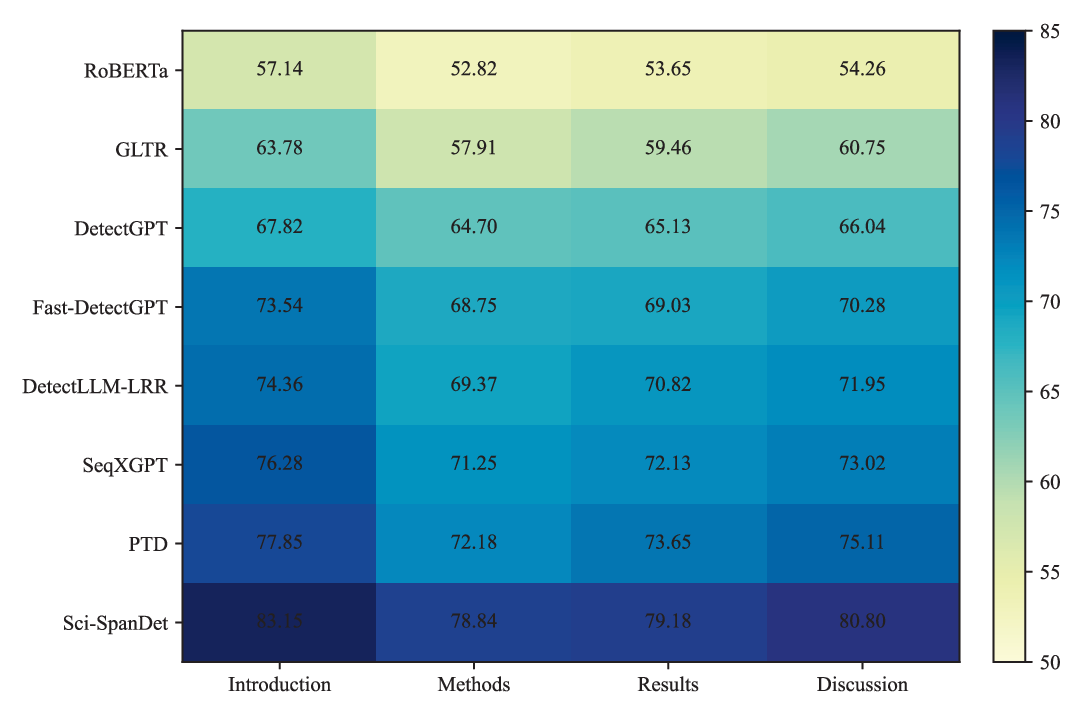}
\vspace{-.5cm}
\caption{Detection performance across structural sections.}
\label{fig:Figure7}
\end{figure}

\noindent\textbf{Section-Level Analysis.} To further analyze the influence of textual structure on detection performance, we evaluate different models on paragraphs sampled from four canonical sections of scientific papers: \textit{Introduction}, \textit{Methods}, \textit{Results}, and \textit{Discussion}. As shown in Fig.~\ref{fig:Figure7}, the detection ability varies substantially across sections due to differences in rhetorical style, terminology density, and discourse function.
Across sections, traditional detectors such as RoBERTa and GLTR yield the lowest F1 scores (55.46 and 61.03 on average), reflecting their reliance on shallow lexical cues. Zero-shot approaches including DetectGPT, Fast-DetectGPT, and DetectLLM-LRR achieve higher performance (above 65 and 70, respectively) but remain unstable, with DetectGPT dropping to 68.75 in the \textit{Methods} section. Span-level baselines further improve robustness: SeqXGPT attains an average F1 of 73.42, and PTD reaches 74.95, benefiting from finer-grained supervision to capture localized rewriting. In comparison, our proposed Sci-SpanDet consistently outperforms all baselines, achieving the highest F1 across sections with 83.15 in \textit{Introduction}, 78.84 in \textit{Methods}, 79.18 in \textit{Results}, and 80.80 in \textit{Discussion}, while also maintaining balanced performance. These results demonstrate that boundary-aware span modeling and section-conditioned contextualization enable Sci-SpanDet to generalize across diverse discourse styles and sustain stable detection accuracy in scholarly text, even in structurally complex sections such as \textit{Methods} and \textit{Results}, thereby ensuring both robustness and practical reliability in real-world academic writing scenarios.

\begin{figure}[t]
    \centering
\vspace{.2cm}    
    \includegraphics[width=1\linewidth]{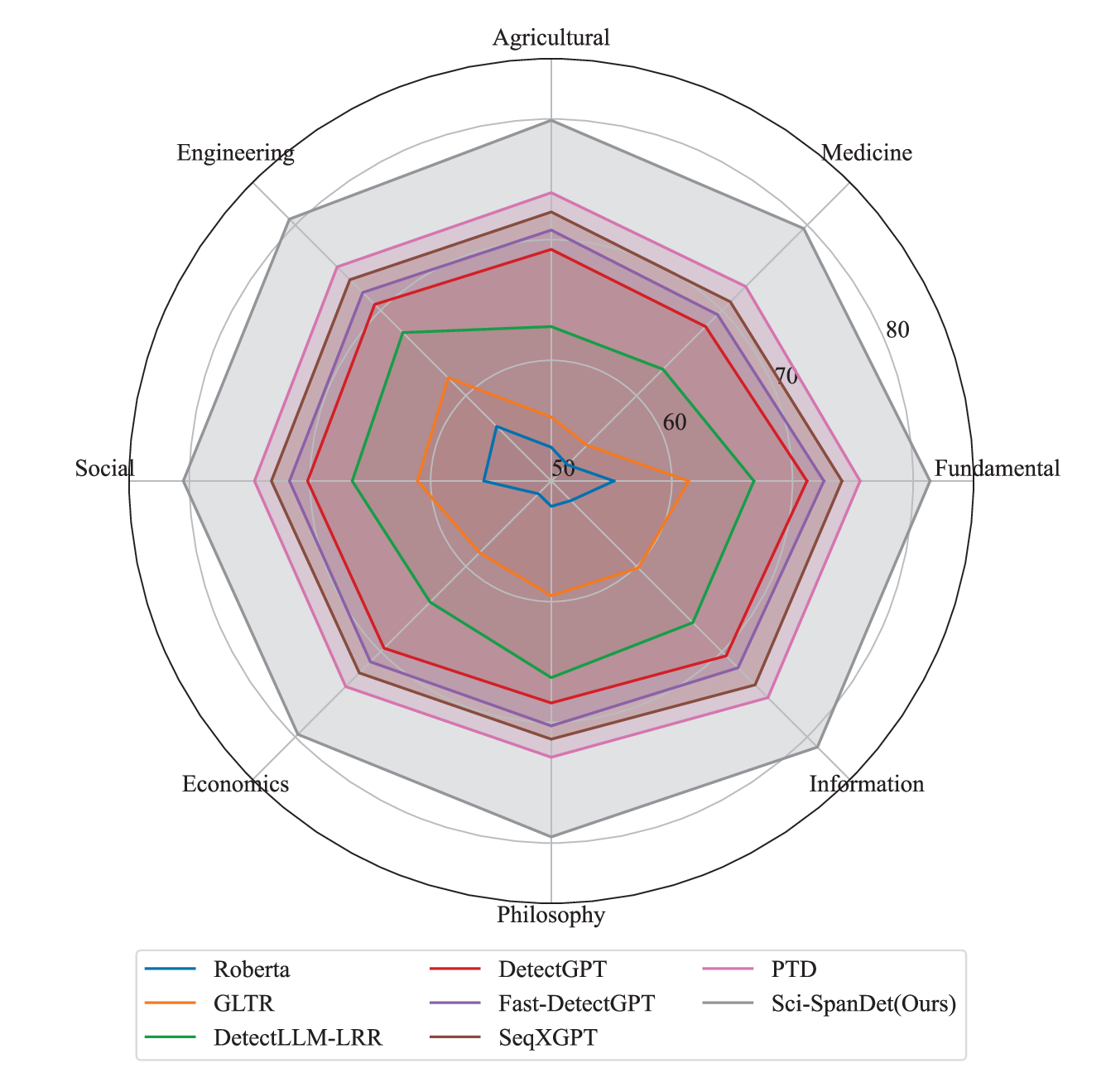}
\caption{Detection performance across academic disciplines.}
\label{fig:Figure8}
\end{figure}

\noindent\textbf{Discipline-Level Analysis.} We further analyze detection performance across eight academic disciplines, including \textit{Fundamental Science}, \textit{Medicine}, \textit{Agricultural Science}, \textit{Engineering}, \textit{Social Science}, \textit{Economics}, \textit{Philosophy}, and \textit{Information Science}. As shown in Fig.~\ref{fig:Figure8}, traditional baselines such as RoBERTa and GLTR achieve relatively low F1 scores (55.46 and 61.03 on average) and display large cross-domain variations, with particularly weak results in \textit{Medicine} (51.9 and 54.2) and \textit{Economics} (51.5 and 58.4). Probability-based zero-shot detectors perform more robustly, as DetectGPT and Fast-DetectGPT maintain F1 values above 68 across all disciplines, while DetectLLM-LRR delivers consistent though moderate gains (average 66.52) through log-rank ratio features. 
Span-level detectors further enhance robustness. SeqXGPT achieves an average of 73.42 with relatively balanced performance (71.0 in \textit{Medicine}, 73.9 in \textit{Information Science}), while PTD pushes the average to 74.95, with notable improvements in \textit{Fundamental Science} (75.6) and \textit{Engineering} (75.1). These results confirm that span-level modeling better captures localized rewriting signals, leading to stronger generalization across domains compared with paragraph-level baselines. In comparison, our proposed Sci-SpanDet consistently delivers the highest scores, ranging from 79.5 in \textit{Philosophy} to 81.4 in \textit{Fundamental Science}. Beyond achieving superior overall accuracy, Sci-SpanDet also exhibits the lowest variance across disciplines, underscoring its adaptability to diverse writing conventions and domain-specific terminologies. This stability is essential for practical deployment in heterogeneous scholarly corpora.

\section{Conclusion}\label{sec:con} 

Sci-SpanDet was proposed as a span-level detection framework tailored to academic texts, integrating section-aware stylistic modeling, graph-based encoding, contrastive learning, and boundary-aware localization with calibrated prediction. Extensive experiments demonstrated that it consistently outperforms existing detection approaches, achieving state-of-the-art performance in F1(AI) and AUROC while uniquely providing accurate span-level localization. The model further exhibited robustness under varying rewriting intensities, balanced performance across paper sections and disciplines, and superior calibration, highlighting both its accuracy and reliability. Overall, Sci-SpanDet advances the detection of AI-generated scientific writing by combining fine-grained interpretability with stable generalization across domains.  

Despite its strengths, our framework relies on accurate section segmentation, and errors in labels may reduce performance since modeling depends on structural cues. Future work will address this limitation, extend to cross-lingual and multimodal data, improve efficiency for large-scale use, and explore identifying the source generator model.

\bibliographystyle{elsarticle-num} 
\bibliography{ref}

\end{document}